\documentclass{article}

    \PassOptionsToPackage{numbers, compress}{natbib}

    \usepackage[preprint]{neurips_2020}

\usepackage{graphicx}

\usepackage[utf8]{inputenc} 
\usepackage[T1]{fontenc}    
\usepackage{hyperref}       
\usepackage{url}            
\usepackage{booktabs}       
\usepackage{amsfonts}       
\usepackage{nicefrac}       
\usepackage{microtype}      
\usepackage{amsmath}
\usepackage{bbm}
\usepackage{wrapfig}
\usepackage[english]{babel}
\usepackage{floatrow}

\usepackage{graphicx}
\usepackage{subcaption}
\usepackage{mwe}

\title{Adaptive Local Bayesian Optimization \\ Over Multiple Discrete Variables}

\author{%
  Taehyeon Kim\thanks{Equally Contributed.} \\
  Graduate School of AI\\
  KAIST\\
  \texttt{potter32@kaist.ac.kr} \\
  \And
  Jaeyeon Ahn$^*$ \\
  Graduate School of AI \\
  KAIST \\
  \texttt{dkswodus49@kaist.ac.kr} \\
  \AND
  Nakyil Kim$^*$ \\
  Graduate School of AI \\
  KAIST \\
  \texttt{nakyilkim@kaist.ac.kr} \\
  \And
  Seyoung Yun \\
  Graduate School of AI \\
  KAIST \\
  \texttt{yunseyoung@gmail.com} \\
}

\begin{document}

\maketitle

\begin{abstract}

In the machine learning algorithms, the choice of the hyperparameter is often an art more than a science, requiring labor-intensive search with expert experience. Therefore, automation on hyperparameter optimization to exclude human intervention is a great appeal, especially for the black-box functions. Recently, there have been increasing demands of solving such concealed tasks for better generalization, though the task-dependent issue is not easy to solve. The Black-Box Optimization challenge (NeurIPS 2020) required competitors to build a robust black-box optimizer across different domains of standard machine learning problems. This paper describes the approach of team KAIST OSI in a step-wise manner, which outperforms the baseline algorithms by up to \textbf{+20.39\%}. We first strengthen the local Bayesian search under the concept of \textit{region reliability}. Then, we design a combinatorial kernel for a Gaussian process kernel. In a similar vein, we combine the methodology of Bayesian and multi-armed bandit\,(MAB) approach to select the values with the consideration of the variable types; the real and integer variables are with Bayesian, while the boolean and categorical variables are with MAB. Empirical evaluations demonstrate that our method outperforms the existing methods across different tasks.


\end{abstract}

\vspace{-10pt}
\section{Introduction}
\vspace{-5pt}
Bayesian optimization\,(BO)\,\citep{shahriari2015taking} has long been known for its sample efficiency without any need for derivative information, and thus it is widely adopted in black-box optimization problems. In each iteration, BO constructs a posterior distribution of the unknown objective (i.e., surrogate model) based on the observations and then selects an expectedly promising point for the next evaluation (i.e., acquisition function). It is particularly known for its performance in hyperparameter optimization within heavily restricted evaluation budgets\,\citep{snoek2012practical}. However, recent literature has addressed that existing works lack consideration on the case when configuration consists of discrete variables\,\citep{ru2019bayesian, moss2020boss}. To deal with this issue, a stream of work has appeared with variations of standard surrogate models and acquisition functions to reflect the different features that discrete variables inherently possess adequately.


The Black-Box Optimization\,(BBO) challenge hosted by NeurIPS\,2020 asked the participants to provide an optimizer capable of tuning a wide range of machine learning problems on real-world datasets\,\footnote[1]{The description about the BBO challenge can be found in \href{https://bbochallenge.com/}{https://bbochallenge.com/}}. Since most of the current studies do not consider the existence of discrete variables, we mainly focused on mitigating the issue of mixed variables. We have applied approaches from different angles to handle the issue.

\begin{figure*}
    \centering
    \includegraphics[width=\textwidth]{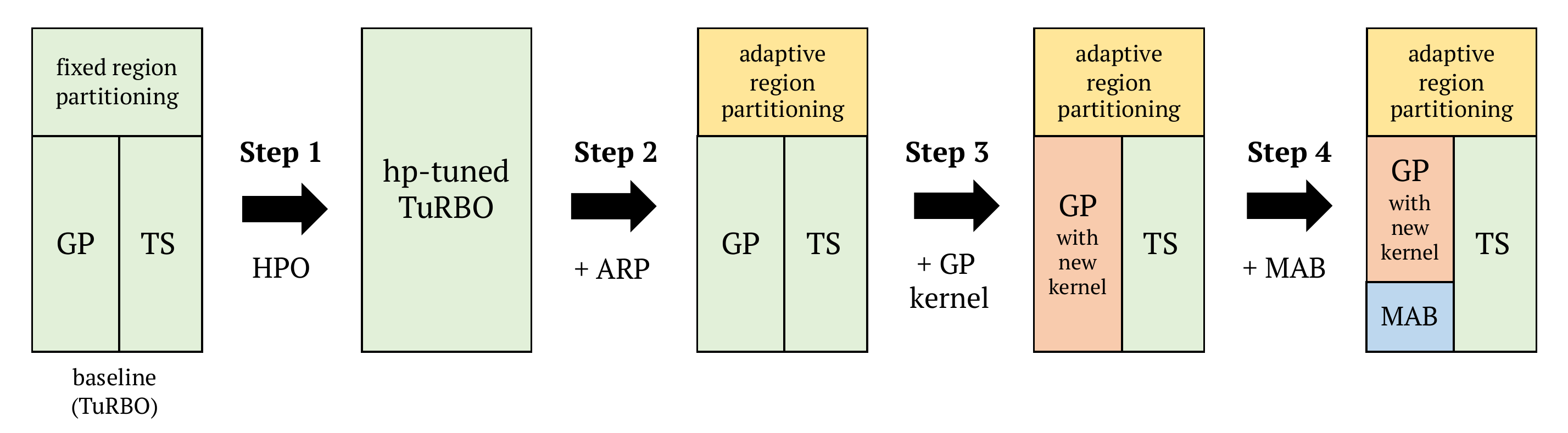}
    \caption[]
    {\small An overview of our algorithm; The baseline consists of local Bayesian modeling which composes a standard Gaussian Process (GP) with mat\`ern kernel as a surrogate model and Thompson sampling (TS) as an acquisition function. After hyperparameter tuning (Step\,1), we compensated for its region partitioning by making it adaptive to different tasks (Step\,2). Then, for the discrete variables, we introduced a mixture of different kernels for GP (Step\,3) and combined MAB for regeneration of qualitative variables (Step\,4).} 
    \label{overall_algorithm}
\end{figure*}

Our solution can be broken down into four steps:\,\textit{(1)} a quick hyperparameter tuning of the baseline model TuRBO (\underline{T}r\underline{u}st \underline{R}egion \underline{BO}), which by itself effectively leads to a global optimum by using a collection of simultaneous \textit{local} optimization with independent probabilistic models \,\citep{eriksson2019scalable}\,(Subsection \ref{turbo section}),\,\textit{(2)} adaptive region partitioning with support vector machine\,(SVM)\,(Subsection \ref{adaptive section}),\,\textit{(3)} combination of different kernels for different types of variables\,(Subsection \ref{kernel section}), and\,\textit{(4)} replacement of qualitative variables (i.e.\,categorical and boolean) by bandit approach based on Thompson sampling\,(Subsection \ref{bandit section}). \autoref{overall_algorithm} illustrates an outline of our algorithm.

\vspace{-5pt}
\subsection{Challenge Overview}
The BBO challenge provided the benchmark package (bayesmark)\,\footnote[2]{Here, the benchmark site is \href{https://github.com/uber/bayesmark}{https://github.com/uber/bayesmark}.}, which supports the local experiments based on the real-world datasets with different objectives and models. The leaderboard score is determined by the performance score on held-out objective functions within the 640 seconds of time limit exclusively on generating suggestions (16 iterations with batch size of 8). The challenge platform provided both the validity and computed score on the validation problem set until the final submission. The official score is decided by the test problem set. 


\vspace{-5pt}
\section{Approaches}

\vspace{-5pt}
\subsection{Baseline: TuRBO} \label{turbo section}
Through the score comparison of the baseline optimizers on the validation set, we chose the TuRBO as our starting point, which ranked the highest performance above all (\autoref{tab:baseline}). 

\vspace{-5pt}
\begin{table}[h]
\centering
\caption{A preliminary survey of the baseline algorithms on the validation problem set.}
\label{tab:baseline}
\resizebox{\textwidth}{!}{%
\begin{tabular}{@{}llllllll@{}}
\toprule
Algorithm & Random Search & HyperOpt\,\citep{bergstra2013making} & SkOpt\,\citep{tim_head_2018_1170575} & SkOpt2\,\citep{tim_head_2018_1170575} & Opentuner\,\citep{ansel:pact:2014} & TuRBO\,\citep{eriksson2019scalable}& PySOT\,\citep{eriksson2019pysot}  \\ \midrule
Score     & 80.224        & 88.127   & 91.867 & 91.932 & 84.801    & \textbf{95.307} & 93.902 \\ \bottomrule
\end{tabular}%
}
\end{table}
\vspace{-5pt}

TuRBO runs multiple bayesian optimizations by incorporating separate subregions which they coined as \textit{trust regions}. Then, it sorts out the next best candidate by implicit Thompson sampling. The algorithm takes standard models for its components where Gaussian process\,(GP) is selected for the surrogate model and Thompson sampling\,(TS) to acquire the next point.


In detail, it first makes a hyper-rectangle centered by the current best observation with its initial length equally set for all dimensions.
The \textit{Sobol} sequence\,\citep{sobol} is sampled from the bounded region, and then it computes their likelihood to select the promising candidates. If the candidate turns out to be the new best point, it perceives success and doubles the length of corresponding trust region. Whereas for all the other cases, it recognizes as a failure and halves them all. If the rescaled length reaches the certain threshold, the trust region restarts from scratch.




We improved the performance of TuRBO through a brief, manual search on hyperparameters that are related to trust region. First, we tuned the minimum length of trust region which directly infers the restarting criterion. In consequence, we settled down with a higher value of minimum length than the default hence the algorithm would behave as more of a quick decision maker for restart (\autoref{tab:minimumlength}). Also, further tuning were made on the number of trust regions for parallel search. TuRBO with multiple trust regions takes the advantage of sampling from separate local models. However, as it consumes a lot of evaluation to initiate individual regions, which is a burden in our constrained setting, we used a single trust region for the subsequent experiments. 


\begin{table}[h]
\centering
\caption{Validation scores according to different minimum length; $2^{-7}$ is the default setting.\label{tab:minimumlength}}
\footnotesize\begin{tabular}{@{}lllllllll@{}}
\toprule
Minimum Length &$ 2^{-1}$ & $2^{-2}$\, & $2^{-3}$\, & $2^{-4}$\, & $2^{-5}$ \, & $2^{-6}$\, & $2^{-7}$\, & $2^{-16}$  \\ \midrule
Score     & 93.694     & 95.186   & \textbf{95.906}   & 95.473 &  95.379 & 95.501  & 95.307 & 93.902 \\ \bottomrule
\end{tabular}%
\end{table}

    

\vspace{-10pt}
\subsection{Adaptive Region Partitioning (ARP)} \label{adaptive section}

Despite the remarkable performance of TuRBO in its quick and efficient optimization, it has several noteworthy weaknesses. First, it owns a stiff criterion on constructing the trust region; i.e., it bounds the local region with fixed rules regardless of the problem setting. Different combinations of objectives and datasets form different landscapes for optimization, hence taking it into account is a better policy. Another weakness is that TuRBO restarts from scratch; therefore time consumption to initiate the new phase may be fatal in a limited budget.





\begin{figure*}[h]
    \centering
        
    \begin{subfigure}[b]{0.18\textwidth}  
        \centering 
        \includegraphics[width=\textwidth]{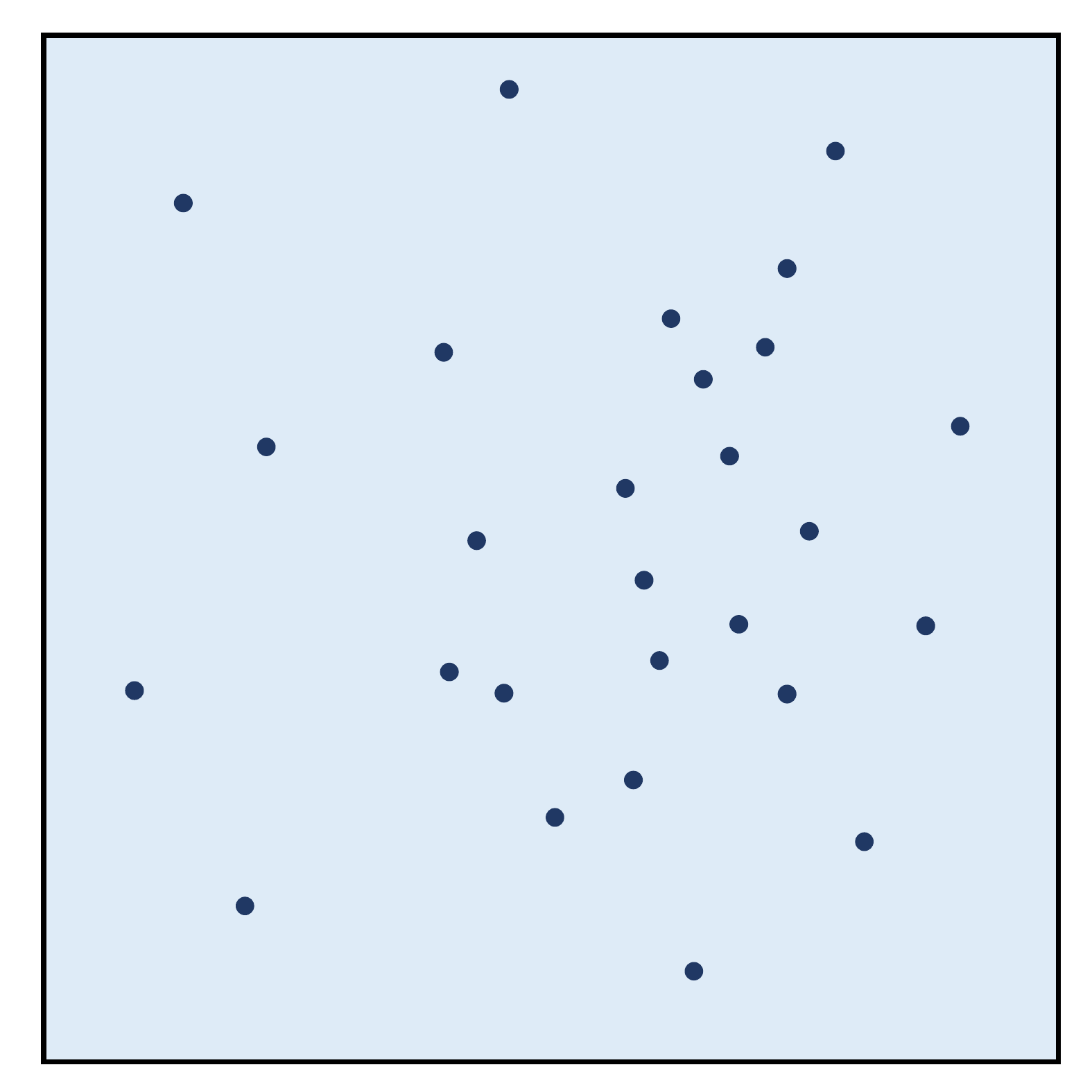}
        \caption{}
        \label{fig2:a}
    \end{subfigure}
    \begin{subfigure}[b]{0.18\textwidth}   
        \centering 
        \includegraphics[width=\textwidth]{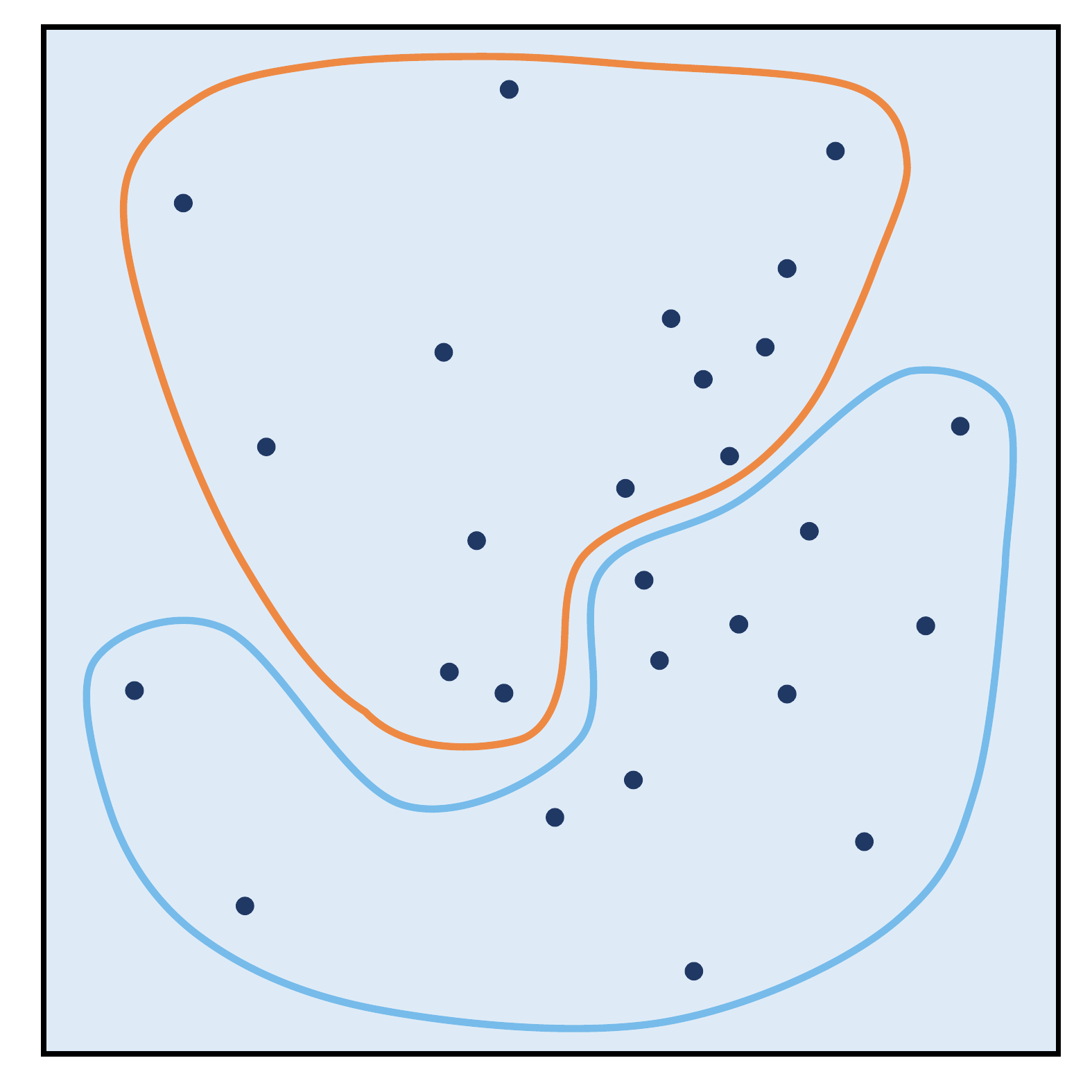}
        \caption{}
        \label{fig2:b}
    \end{subfigure}
    \begin{subfigure}[b]{0.18\textwidth}   
        \centering 
        \includegraphics[width=\textwidth]{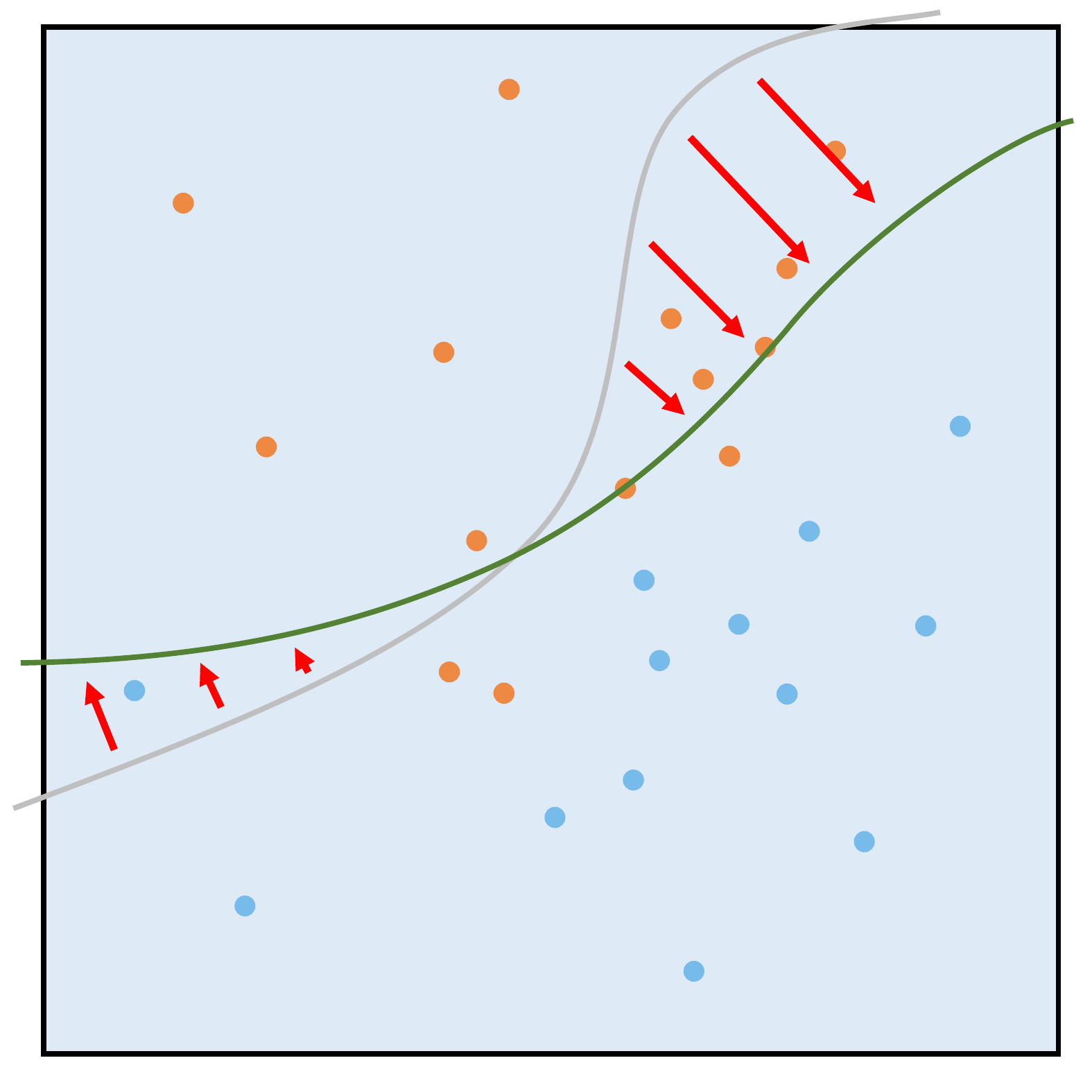}
        \caption{}
        \label{fig2:c}
    \end{subfigure}
    \begin{subfigure}[b]{0.18\textwidth}   
        \centering 
        \includegraphics[width=\textwidth]{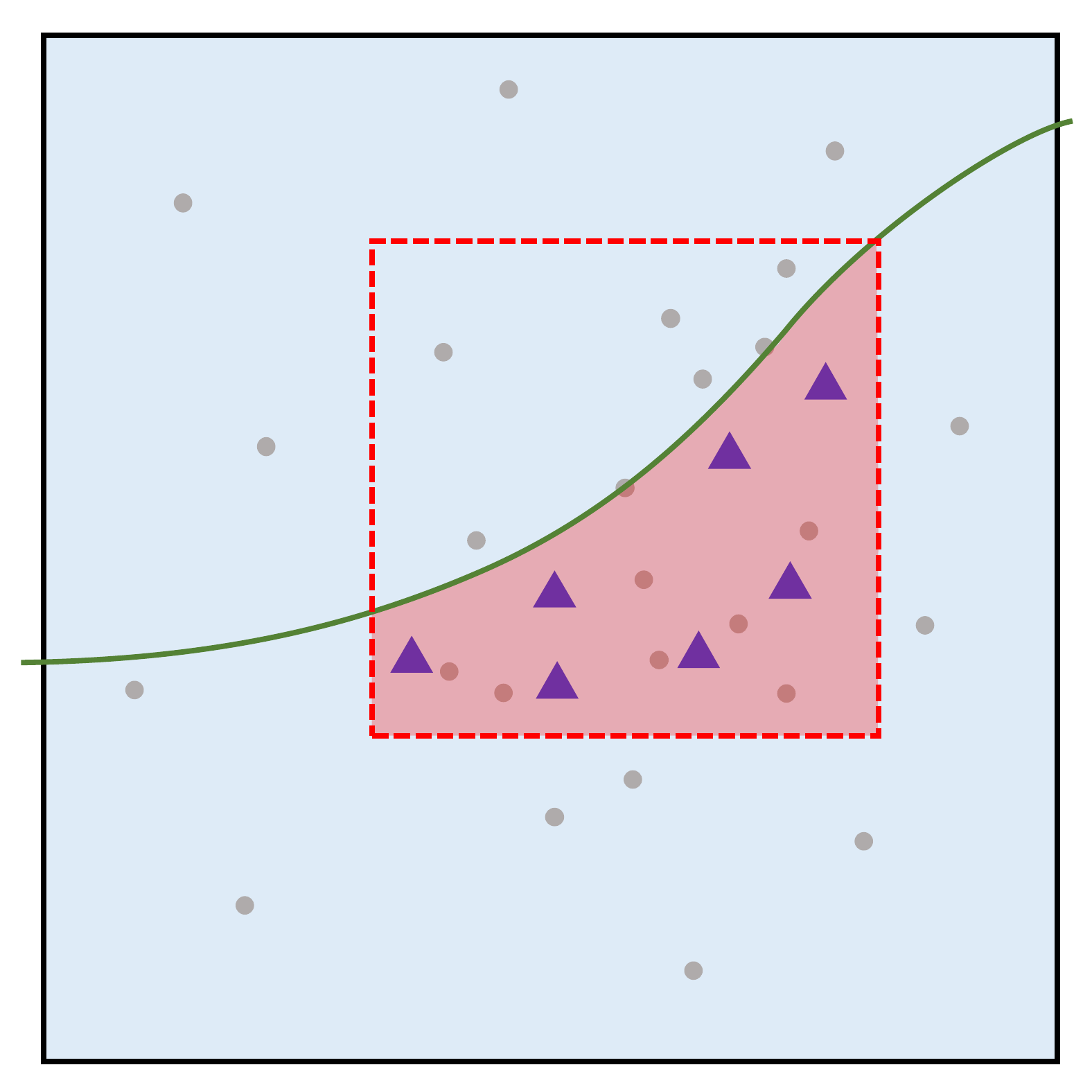}
        \caption{}
        \label{fig2:d}
    \end{subfigure}
    \begin{subfigure}[b]{0.20\textwidth}   
        \centering 
        \includegraphics[width=\textwidth]{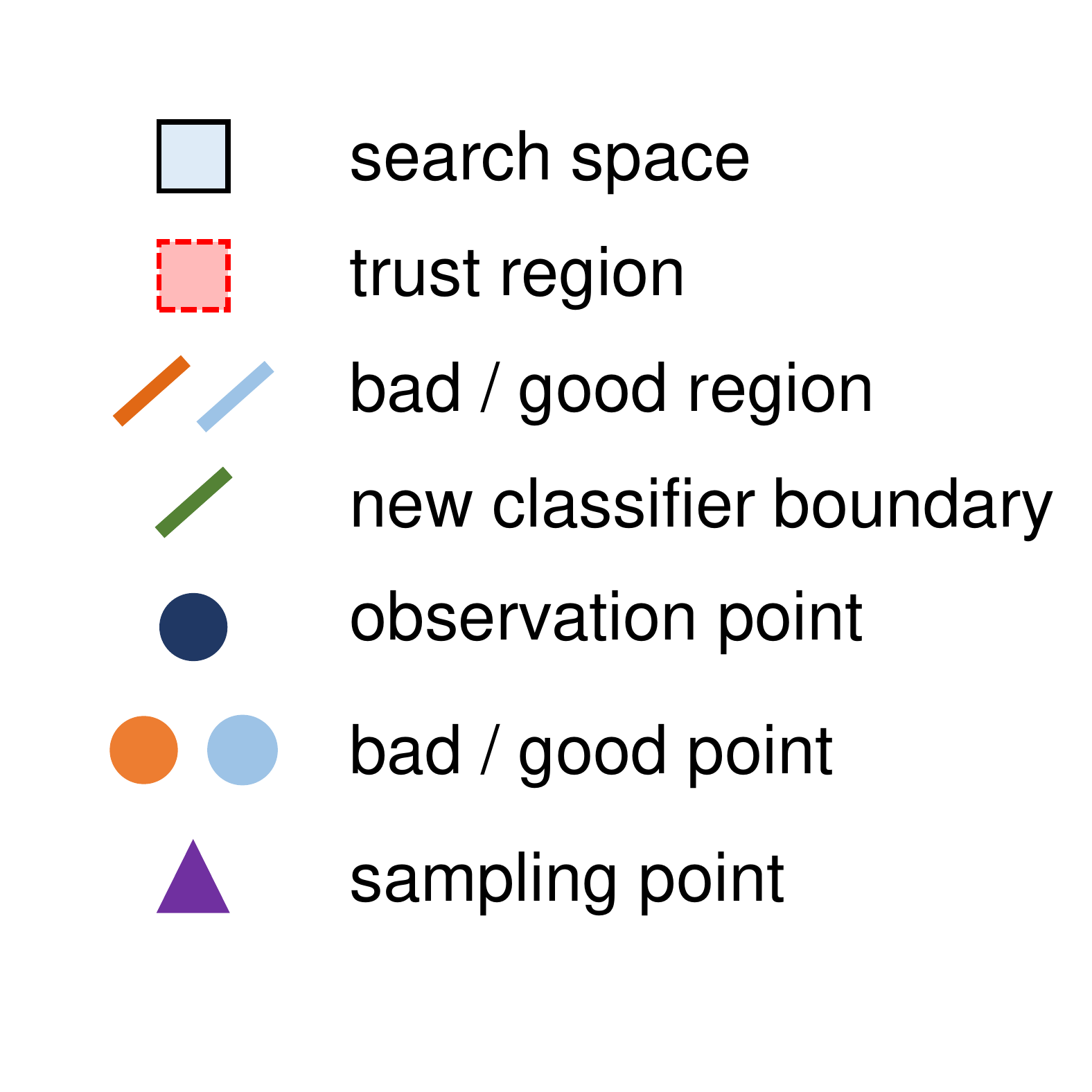}
        \label{fig:arp_labels}
    \end{subfigure}

    \caption{\small An overview of ARP; (\subref{fig2:a}) the dots represent the evaluated observations in the search space, (\subref{fig2:b}) first, we produce labels with k-means clustering, (\subref{fig2:c}) then, we retrain the classifier to learn the new boundary, (\subref{fig2:d}) the trained classifier is used to partition the trust region so that the next sampling is generated within the selected region. Here we select the region that includes the current best point. The clustering and classifier training are held under global observations.} 
    \label{arp_algorithm}
    
\end{figure*}

To alleviate these drawbacks, we proposed a simple yet effective method, called ARP\,(\underline{A}daptive \underline{R}egion \underline{P}artitioning) with a hint from \textit{Wang et al.}\,\citep{wang2020learning}. ARP is a method of adding a classifier as a helper tool not only to reshape the constructed trust region but also to restart within a promising region\,(\autoref{fig2:d}). ARP is activated after stacking a sufficient amount of observations. First, at each iteration, the observations are labeled into two groups - good region and bad region - based on the evaluation values by k-means clustering\,(\autoref{fig2:b}). Then, SVM learns the decision boundary through synthetic labels\,(\autoref{fig2:c}), and as a consequence, it can be used to partition within the current trust region so that candidates are extracted from the selected\,(good) region\,(\autoref{fig2:d}). Also in the restart phase, random samples are created within the selected region using the pretrained classifier. Then, even after a number of restarts, the newly initiated trust region can be built upon a reliable subregion instead of the whole search space.

Our key differentiation from \textit{Wang et al.}\,\citep{wang2020learning} is that our algorithm utilizes all the data points to train the classifier, i.e. the partition always consider the entire region. In \textit{Wang et al.}\,\citep{wang2020learning}, they bifurcate the search space into a tree structure and then train the classifier only with the points within the selected local region. While their strategy may be effective when sufficient amount of budget is given, it would be a bottleneck under the consideration of the limited iterations of the challenge. In this respect, the entire data is employed to learn the region partitioning in our implementation.


\vspace{-5pt}
\subsection{A Mixture of Different GP Kernels} \label{kernel section}


In this subsection, we introduced a new design of kernel to remedy the absence of the consideration of discrete variables. Recent studies have suggested a solution of mixing multiple kernels\,\citep{kandasamy2015high, ru2019bayesian}. For instance, a summation of additive kernels\,\citep{kandasamy2015high} and a combination of sum and product of two separate kernels for continuous inputs and categorical inputs\,\citep{ru2019bayesian} are discussed. From the insight of the previous works\,\citep{kandasamy2015high, ru2019bayesian}, we adopted a novel kernel as follows:

\vspace{-10pt}
\begin{equation}\label{eqone}
        {k}_{M}(\mathbf{x}, \mathbf{x^{'}}) = \frac{2^{1- \nu }}{\Gamma{(\nu)}} (\sqrt{2\nu}d) {K_{\nu}} (\sqrt{2\nu}d), \quad
    {k}_{L}(\mathbf{y}, \mathbf{y^{'}}) = {v}\mathbf{y^{T}} \mathbf{y^{'}}, \quad 
    {k}_{I}(\mathbf{z}, \mathbf{z^{'}}) = \mathbbm{1}(\mathbf{z}, \mathbf{z^{'}}) 
\end{equation} 
\vspace{-10pt}
\begin{equation} \label{eqtwo}
    {k}(\mathbf{h}, \mathbf{h^{'}}) = (1-\lambda)  ({k}_{M}(\mathbf{x}, \mathbf{x^{'}}) + {k}_{L}(\mathbf{y}, \mathbf{y^{'}}) + {k}_{I}(\mathbf{z}, \mathbf{z^{'}})) + \lambda {k}_{M}(\mathbf{x}, \mathbf{x^{'}}){k}_{L}(\mathbf{y}, \mathbf{y^{'}}){k}_{I}(\mathbf{z}, \mathbf{z^{'}}) 
\end{equation}


where $h$ is an input that decomposed into $x$, $y$, and $z$, each respectively matching to continuous, integer, and categorical\,/\,boolean variables, $d$ indicates the Euclidean distance between $x$ and $x^{'}$,  $\lambda$ is the balancing parameter between kernels, $ \nu $ serves as the smoothness parameter for mat\`ern kernel, $ v $ represents the signal variance parameter for linear kernel, and $\mathbbm{1}(z, z^{'})=1$ if $z=z^{'}$ and otherwise 0. For the implementation, we fixed the value as 2.5 for $\nu$ and 1 for $ v $, which generally worked well in our local experiments. 

Separating kernels would detect different inherent features of each type, where the specific type of kernel is chosen based on the empirical result on the validation set. As shown in Eq.\,\eqref{eqone} and \eqref{eqtwo}, these kernels are integrated by linear combination of product and sum to learn the correlation between types of variables. 

\vspace{-5pt}
\subsection{Multi-Armed Bandits on Qualitative Variables} \label{bandit section}

\begin{wrapfigure}[16]{r}{0.4 \linewidth}
    \vspace{-15pt}
    \includegraphics[width=\linewidth]{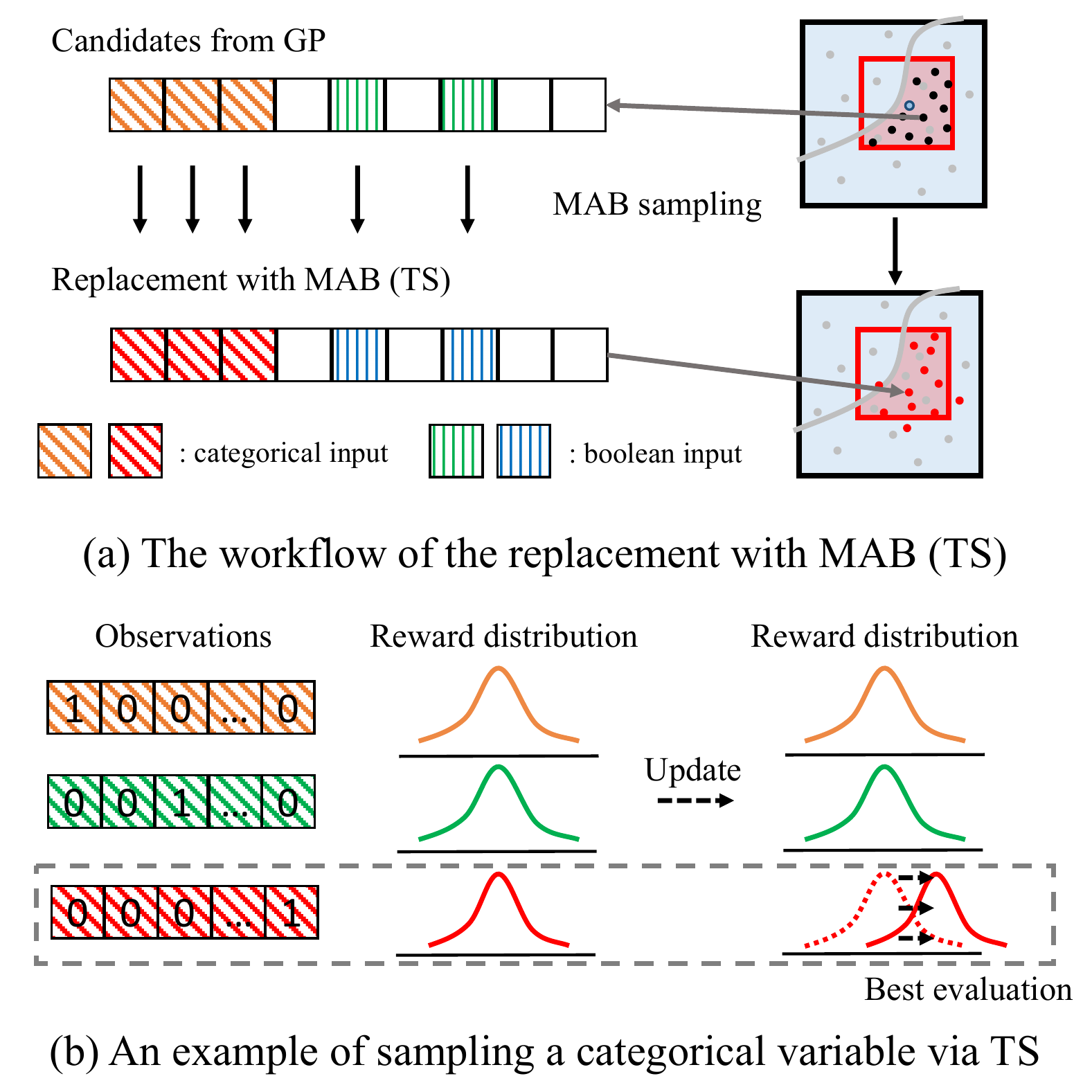}
    \begin{center}
    \vspace{-25pt}
    \caption{An overview of bandit. \label{TS_overview}}
    \end{center}
\end{wrapfigure}



After extracting candidates from the GP (Subsection \ref{kernel section}), the values of categorical and boolean variables are replaced with the samples from Thompson sampling (TS). 
We empirically found out that TuRBO's local search strategy led to repeated bounding around same points, thereby particularly providing excessive exploitation on certain types of variables. In detail, TuRBO often seems to select the same values for the qualitative variables repeatedly during one phase of trust region search. TuRBO's shrinking policy on failure intensifies this lack of exploration; as a result, it would inevitably spend redundant time in limited space where the bounded region may be worthless for further search.




To compensate the insufficient exploration of local search, we add an alternative for the aforementioned variables that are particularly under-explored. In specific, we chose MAB as a substitute to generate categorical and boolean variables while leaving float and integer with GP. Although we made a partial replacement with MAB, the GP with our constructed kernel enables the algorithm to learn the correlation between different variables. For integer variable, despite its discreteness, we exclude it from applying MAB since it rather acts as a continuous variable in the sense that both are quantitative \,\citep{swiler2014surrogate, pelamatti2020overview, daxberger2019mixed}. If the candidate turns out to be the best evaluation, each $\alpha$ parameter in the beta distribution of the corresponding arms increases\,\footnote[4]{The beta distribution of each arm is initialized as $Beta\,(\alpha=1, \beta=1)$.}\,(\autoref{TS_overview}\,(b)) as a reward.

\vspace{-10pt}
\section{Conclusion and Future works}
\vspace{-5pt}

\begin{wraptable}{r}{7.5cm}
\vspace{-10pt}
\centering
\ttabbox{
\resizebox{\textwidth}{!}{%
\begin{tabular}{@{}c|ccccc@{}}
\toprule
Algorithm & Baseline & + Tuning & + ARP & + Mixture Kernel \& Bandit  \\ \midrule
Score     & 95.307  & 95.906  & 96.298    & \textbf{96.580} \\ \bottomrule
\end{tabular}%
}}{\caption{The validation scores of the algorithm; the preceding additions are included in each steps}\label{tab:submitted_result}
}
\end{wraptable}

Along with the described steps, we have recorded the leaderboard scores, which clearly show that each addition enhanced the baseline to a certain extent (\autoref{tab:submitted_result}). Regarding the final evaluation on the test problem set, our method rated \textbf{90.875} as an official score, ranking 8th in the final leaderboard.

We have mainly focused on treating the discrete variables, which distinctively possess different characteristics comparing to continuous variables. Our approach numerically proved its improvement comparing to the baseline methods (see \autoref{tab:impro} in \autoref{appendix:b}). Despite its superiority, there are still lacks on the theoretical analysis about the regret bound of MAB and the convergence guarantee of our designed kernel. As future work, a theoretical analysis of the remarkable performance of local search can be expected.





\begin{ack}
This work was supported by Institute for Information \& Communications Technology Promotion (IITP) grant funded by the Korea government (MSIT) [No.2018-0-00278, Development of Big Data Edge Analytics SW Technology for Load Balancing and Active Timely Response] and [No.2019-0-00075, Artificial Intelligence Graduate School Program(KAIST)]. We thank Jonghyup Kim for insightful discussion about this challenge.
\end{ack}

\bibliography{main}
\bibliographystyle{plain}
\appendix
\section{Challenge Overview}
\begin{figure}[h]
     \centering
         \includegraphics[width=\textwidth]{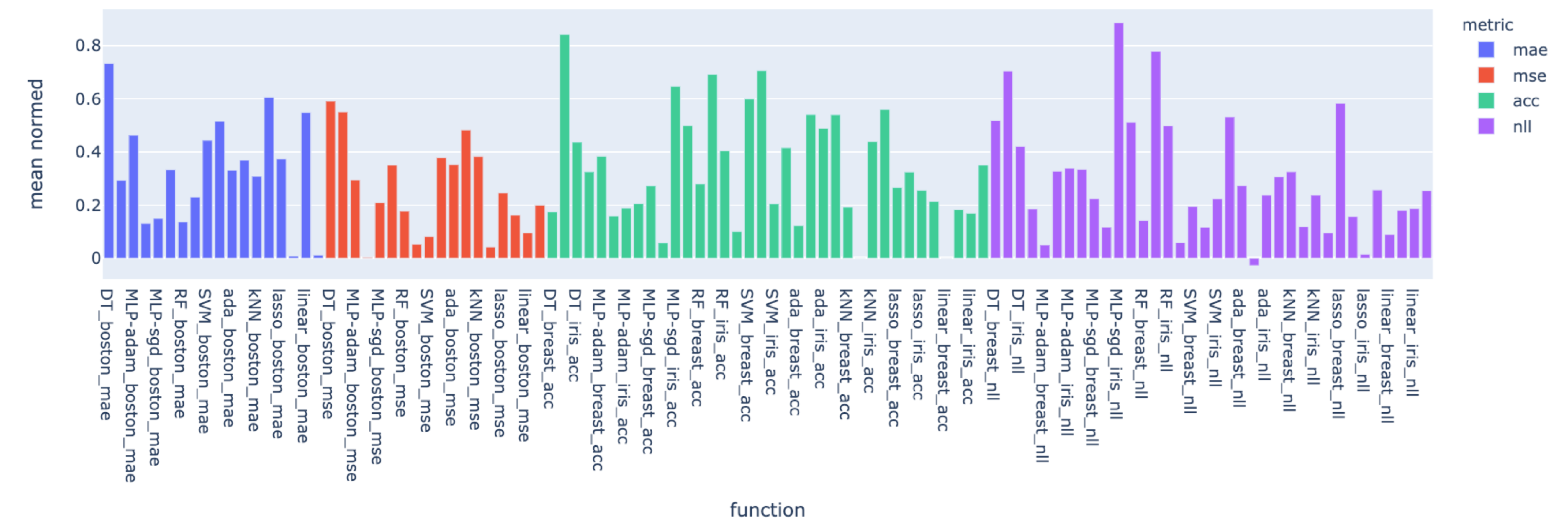}
     \caption{A result of the TuRBO\,\citep{eriksson2019scalable} on the local experiments in the benchmark package. Under the given scikit-learn datasets, a black-box optimizer is evaluated with various combinations of machine learning models\,(e.g. MLP, linear, kNN) and metrics\,(e.g. mae, mse, acc, nll).}
    \label{fig:local_ex}
\end{figure}

The training datasets can be found in \href{https://scikit-learn.org/stable/datasets/index.html\#toy-datasets}{https://scikit-learn.org/stable/datasets/index.html\#toy-datasets}\,(\autoref{fig:local_ex}). The leaderboard is determined using the optimization performance on held-out (hidden) objective functions, where the optimzier must run without human intervention.

\section{Improvement of Our Algorithm on Validation Set}
\label{appendix:b}
\vspace{-5pt}
\begin{table}[h]
\centering
\caption{A preliminary survey of the baseline algorithms on the validation dataset. Improvement was calculated based on equation $ \frac{(S_o - S_b )}{S_b}$ where $S_o$ and  $S_b$ indicates our score and baseline score, respectively.}
\label{tab:impro}
\resizebox{\textwidth}{!}{%
\begin{tabular}{@{}lllllllll@{}}
\toprule
Algorithm & Random Search & HyperOpt\,\citep{bergstra2013making} & SkOpt\,\citep{tim_head_2018_1170575} & SkOpt2\,\citep{tim_head_2018_1170575} & Opentuner\,\citep{ansel:pact:2014} & TuRBO\,\citep{eriksson2019scalable}& PySOT\,\citep{eriksson2019pysot} & Ours  \\ \midrule
Score     & 80.224        & 88.127   & 91.867 & 91.932 & 84.801    & 95.307 & 93.902 & \textbf{96.580} \\ \midrule
Improvement & + 20.39\% & + 9.59 \% & + 5.13 \% & + 5.06 \% & + 13.89 \% & + 1.34 \% & + 2.85 \% & - \\ \bottomrule
\end{tabular}%
}
\end{table}
\vspace{-5pt}

\end{document}